\documentclass[sigconf]{aamas}

\usepackage{multirow}
\usepackage{balance} 

\usepackage{enumitem}

\doi{EFXQ8322}

\makeatletter
\gdef\@copyrightpermission{
  \begin{minipage}{0.2\columnwidth}
   \href{https://creativecommons.org/licenses/by/4.0/}{\includegraphics[width=0.90\textwidth]{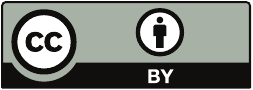}}
  \end{minipage}\hfill
  \begin{minipage}{0.8\columnwidth}
   \href{https://creativecommons.org/licenses/by/4.0/}{This work is licensed under a Creative Commons Attribution International 4.0 License.}
  \end{minipage}
  \vspace{5pt}
}
\makeatother

\setcopyright{ifaamas}
\acmConference[AAMAS '26]{Proc.\@ of the 25th International Conference
on Autonomous Agents and Multiagent Systems (AAMAS 2026)}{May 25 -- 29, 2026}
{Paphos, Cyprus}{C.~Amato, L.~Dennis, V.~Mascardi, J.~Thangarajah (eds.)}
\copyrightyear{2026}
\acmYear{2026}
\acmDOI{}
\acmPrice{}
\acmISBN{}

\title[AAMAS-2026 Formatting Instructions]{Synthesis and Evaluation of Long-term History-aware Medical Dialogue}

\author{Hebin Hu}
\authornote{The first two authors contributed equally to this work.}
\affiliation{
  \institution{South-Central Minzu University}
  \city{Wuhan}
  \country{China}}
\email{hebin9410@gmail.com}

\author{Renke Dai}
\authornotemark[1]
\affiliation{
  \institution{South-Central Minzu University}
  \city{Wuhan}
  \country{China}}
\email{2024110293@mail.scuec.edu.cn}

\author{Ah-Hwee Tan}
\authornote{Corresponding author}
\affiliation{
  \institution{Singapore Management University}
  \city{Singapore}
  \country{Singapore}}
\email{ahtan@smu.edu.sg}

\author{Yilin Kang}
\authornotemark[2]
\affiliation{
  \institution{South-Central Minzu University}
  \city{Wuhan}
  \country{China}}
\email{ylkang@mail.scuec.edu.cn}

\begin{abstract}

An effective healthcare agent must be able to recall and reason over a patient’s longitudinal medical history. However, the absence of datasets with realistic long-term dialogue timelines limits systematic evaluation. Real clinical text is constrained by privacy and ethics, while existing benchmarks focus on isolated interactions, failing to capture cross-session reasoning. We introduce a framework for synthesizing high-quality, long-term medical dialogues with LLMs. Our approach entails a knowledge-guided decomposition into three stages: constructing synthetic patient profiles with diverse disease and complication trajectories, generating multi-turn dialogues per encounter, and integrating them into a coherent longitudinal history dataset, MediLongChat. We establish three benchmark tasks—In-dialogue Reasoning, Cross-dialogue Reasoning, and Synthesis Reasoning—to evaluate the memory capabilities of healthcare agents. To assess data quality, we introduce a multi-dimensional evaluation framework combining vector-based metrics with LLM-as-a-judge assessments. Specifically, we define automatic measures—Faithfulness, Coherence, and Diversity—together with two LLM-based evaluations: Correctness and Realism. Benchmark experiments show that even state-of-the-art LLMs struggle with MediLongChat. These findings highlight the benchmark’s applicability and underscore the need for tailored methods to advance healthcare agents.

\end{abstract}


\newcommand{\BibTeX}{\rm B\kern-.05em{\sc i\kern-.025em b}\kern-.08em\TeX}

\begin{document}


\pagestyle{fancy}
\fancyhead{}


\maketitle 


\section{Introduction}
\begin{figure}
    \centering
    \includegraphics[width=0.95\linewidth]{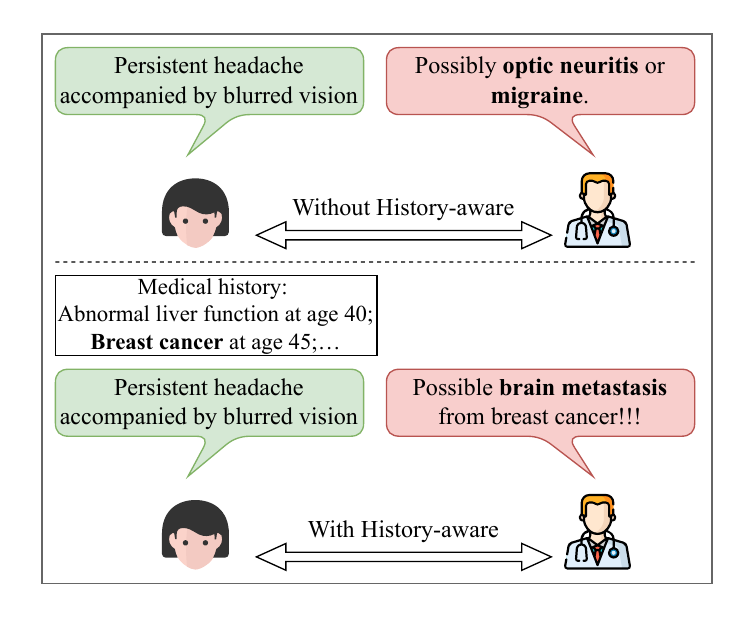}
    \caption{The importance of history-aware capabilities for healthcare agents. A history-agnostic model defaults to common causes for “headache + blurred vision,” whereas a history-aware model recalls prior breast cancer, yielding different diagnoses and safety actions.}
    \label{fig: History-aware}
\end{figure}

 As large language models (LLMs) have offered significant assistance in diverse areas in the medical domain \cite{medrag, xu2019end, singhal2023large}, a central challenge lies in the development of healthcare agents that can engage in long-term, coherent conversations. A core requirement for such agents is the ability to interpret and utilize longitudinal patient history—not just the current utterance, but months or years of interactions concerning prior symptoms, diagnostics, and treatments. We refer to these settings as history-aware longitudinal clinical dialogues. For instance, as shown in Fig.~\ref{fig: History-aware}, consider a patient who now reports persistent headaches and blurred vision. A history-agnostic agent might default to migraine. In contrast, a history-aware agent that recalls a prior breast-cancer diagnosis and incomplete follow-up imaging will surface altogether different diagnoses and safety actions. This example illustrates a significant point: Synthesize Longitudinal Reasoning-drawing clinically valid inferences from events scattered along a longitudinal trajectory—is fundamental to building safe and reliable healthcare agents.

Despite this importance, existing public benchmarks rarely stress longitudinality. Popular dialogue corpora tend to comprise independent conversations lacking a consistent patient narrative, while medical QA benchmarks emphasize static knowledge rather than dynamic, context-dependent reasoning \cite{jin2019pubmedqa, pal2022medmcqa}. Moreover, assembling real longitudinal dialogue data is ethically and operationally daunting: clinical text is privacy-sensitive; de-identification is costly and imperfect; and even approved corpora frequently cover narrow settings and come with stringent governance. As a result, research on long-horizon clinical dialogue remains hampered by data scarcity.

Although the synthetic approach alleviates data scarcity and compliance barriers \cite{kovavcevic2024identification, gonzales2023synthetic}, its quality is constrained by three common bottlenecks:
\begin{itemize}[leftmargin=*, align=left]
    \item Generation Quality and Consistency. The inherent hallucination problem \cite{li2023halueval, manakul2023selfcheckgpt} of LLMs is particularly detrimental in the medical field. This issue is compounded by their tendency to produce contradictory details when generating long dialogues 
    \cite{spataru2024know, liu2024lost}. Furthermore, the Mixture of Experts (MoE) architecture used in many LLMs may exhibit instability when generating long texts, leading to inconsistencies in style or knowledge depth.

    \item Context Window Limitations. 
    A fundamental barrier to generating long, continuous dialogues is the finite context window of LLMs. While nowadays models support longer contexts, utilizing their full capacity is often prohibitively expensive and still falls short of capturing a complete patient history.This exposes a deeper, unsolved problem: how to architect generation processes—from single, guided passes to complex pipelines—to guarantee narrative and logical coherence over the long term.

    \item Lack of Evaluation Standards. There is a lack of standardized methods for evaluating the quality of synthetic data itself. Current research often relies on limited automatic metrics or small-scale human evaluations, lacking a systematic and scalable framework to comprehensively measure a dataset's medical accuracy, long-term logical consistency, and effectiveness for evaluating agent capabilities.
    
\end{itemize}

To address the aforementioned challenges, this paper proposes a systematic pipeline for generating history-aware longitudinal clinical dialogue datasets and a multi-dimensional framework for their evaluation. The core of our method lies in task decomposition guided by structured knowledge. Specifically, we first construct metadata about disease cases and their complications. Based on this, we generate synthetic patient profiles with complete, diverse, and chronologically ordered medical events. Our task decomposition approach breaks down the complex process of generating a patient's lifelong medical history into manageable steps. By progressively creating multi-turn dialogues for each clinical visit, we build a long-term, coherent, and history-aware medical conversation dataset MediLongChat. 

To systematically evaluate the memory and reasoning abilities of healthcare assistants, we introduce a benchmark built upon our dataset, featuring three dedicated tasks: In-dialogue Reasoning, Cross-dialogue Reasoning, and Synthesis Reasoning. These tasks respectively assess an agent's ability to recall information from a single encounter, link events across multiple dialogues, and synthesize the complete history for clinical inference. 
More importantly, we have designed a comprehensive framework that combines automatic metrics with an LLM-as-a-Judge approach to evaluate the quality of the generated dataset by measuring its Faithfulness, Coherence, Correctness, Diversity, and Realism. 

The main contributions of this paper are summarized as follows:
\begin{itemize}[leftmargin=*, align=left]
    \item We propose a novel framework for synthesizing long-form, history-aware medical dialogues with explicit longitudinal dependencies, based on knowledge-guided task decomposition, directly addressing challenges of LLM hallucination and inconsistency in long-content generation.
    \item We propose a comprehensive evaluation framework that establishes a new standard for assessing the quality of synthetic medical dialogue data.
    \item We construct a new benchmark dataset, MediLongChat, specifically designed to evaluate the longitudinal memory and reasoning capabilities of healthcare agents in multi-session dialogues.
\end{itemize}


\section{Related Work}
\subsection{Synthetic Medical Datasets}
In the medical field, growing concerns over privacy, ethics, and data scarcity have led researchers to increasingly favor synthetic or semi-synthetic data for model training and evaluation. NoteChat \cite{NoteChat} generates doctor-patient dialogues from clinical notes using a multi-agent framework, incorporating medical logic control to minimize invalid outputs. SynDial \cite{SynDial} leverages publicly available MTS-Dialogue and MIMIC datasets to generate dialogues via zero-shot prompting, integrating a feedback loop during generation to enhance dialogue quality. This method demonstrates superior performance in extractive and factual consistency compared to simple prompting approaches. 
SynSUM \cite{synsum} bridges structured variables and clinical text for information extraction and causal studies, employing a Bayesian network to first generate tabular variables, which are then used to prompt an LLM to produce corresponding clinical text. Holysz et al. propose a multi-stage generation framework that first generates patient profiles and case backgrounds before producing dialogues, striving to improve the diversity and medical plausibility of synthetic data.

Although the aforementioned methods can generate high-quality single-session dialogues or dialogue-note paired samples, they still exhibit typical limitations in cross-turn and cross-dialogue reasoning. Most existing datasets focus on single consultation dialogues or intra-session dialogue-note alignment, with very few encompassing longitudinal records or dialogue histories of the same patient across multiple sessions. This presents a significant gap in evaluating whether healthcare agents possess long-term memory or the ability to comprehend longitudinal patient history.

\subsection{Dataset Evaluation}

\label{sec:related_work}

Evaluating the quality of dialogue datasets and models is a critical and long-standing challenge in the field of natural language processing. Prior work can be broadly categorized into three main approaches: human-centric evaluation, traditional lexical and vector-based methods, and more recent LLM based evaluation.

Human Evaluation remains the gold standard for assessing dialogue quality \cite{deriu2021survey}. For single-turn dialogues, metrics like mean opinion score or pairwise comparisons are widely used \cite{deriu2021survey,li2016deep}. However, evaluating multi-turn conversations presents unique challenges. Annotators must consider the entire dialogue history to assess long-term coherence, consistency, and the model's ability to maintain a persona or follow a complex narrative arc. While highly reliable, human evaluation is expensive, time-consuming, and difficult to scale, especially for large datasets with thousands or millions of dialogues. Moreover, achieving high inter-annotator agreement can be challenging due to the subjective nature of dialogue quality.

Traditional Lexical and Vector-Based Metrics were introduced to address the scalability issues of human evaluation. Early approaches focused on lexical overlap. Metrics such as BLEU \cite{papineni2002bleu}, ROUGE \cite{lin2004rouge}, and METEOR \cite{banerjee2005meteor} measure the n-gram similarity between a model's response and a set of human-written reference responses. While effective for tasks with a limited range of correct answers, these metrics are less suitable for multi-session dialogue, where there can be many valid and novel responses that do not match the reference text. To capture semantic similarity, later methods employed word embeddings and other vector-based techniques. Metrics like embedding average or greedy matching \cite{rush-etal-2015-greedy} calculate the cosine similarity between the embeddings of the generated response and the reference response. These metrics offer an improvement over lexical overlap by accounting for synonyms and semantically similar words. However, they still struggle to evaluate nuanced aspects of dialogue quality, such as factual consistency, long-term coherence across multiple turns, and the overall conversational flow of a lengthy dialogue.


Recently, the remarkable capabilities of LLMs have led to their use as automated evaluators, often referred to as "LLM-as-a-judge" \cite{zheng2023judging}. The LLM is given a dialogue and a set of instructions, and it returns a score, a ranking, or a detailed critique. This method offers several advantages: it is significantly faster and cheaper than human evaluation, and it can be designed to assess more complex attributes like nuance, factual accuracy, and conversational flow. However, LLM-based evaluation is not without its limitations. The judgments can be susceptible to biases, such as position bias \cite{zheng2023judging} or verbosity bias. While LLM-as-a-judge has shown high correlation with human judgments in many cases, it is still a developing field, and the reliability and robustness of these methods for complex, multi-turn conversations remain an active area of research.

\begin{figure*}[!htbp]
    \centering
    \includegraphics[width=1\textwidth]{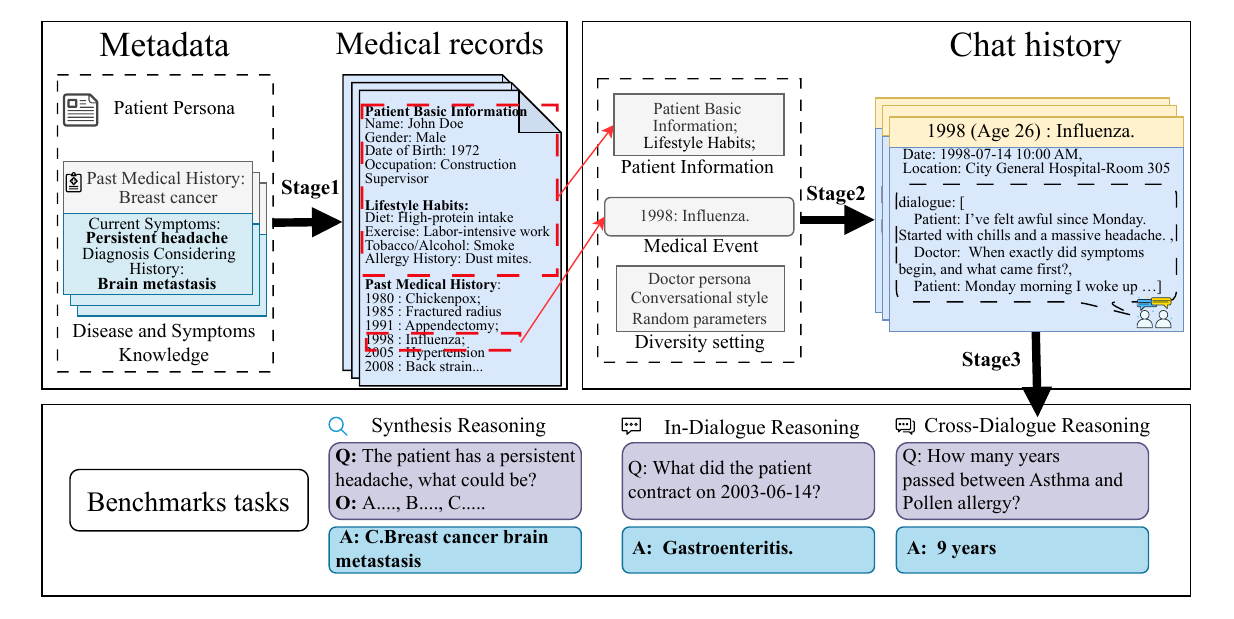}
    \caption{Overview of our dataset generation pipeline. Stage 1 builds records with knowledge guidance; Stage 2 generates per-event encounters and stitches them chronologically; Stage 3 derives three tasks to assess longitudinal memory and reasoning.
    }
    \label{fig: dataset generation}
\end{figure*}

\section{MediLongChat Synthesis Pipeline}
Our goal is to construct a high-quality, long-sequence medical dialogue dataset together with a benchmark that evaluates long-range memory and reasoning in healthcare agents. As illustrated in Fig.~\ref{fig: dataset generation}, the framework comprises three stages: (1)Knowledge-Guided Generation of Patient Medical Records; (2)Multi-turn Dialogue Generation Based on Task Decomposition; (3)Benchmark Generation.

\subsection{Stage 1: Knowledge-Guided Generation of Patient Medical Records}
A medical record contains complete information about a specific patient. Each medical record consists of four components: personal information, lifestyle habits, past medical history, and additional information. Medical records represent fictitious yet realistic lifelong patient profiles, thus avoiding privacy concerns. Notably, medical records serve as intermediate data during the generation process and are not included explicitly in the final dataset. Also, they remain hidden from the LLMs during evaluation benchmarks. High-quality datasets start from high-quality priors. To mitigate medically incorrect hallucinations, we generate complete fictional patient records under explicit knowledge guidance in three steps, ensuring both medical plausibility and narrative diversity.

\subsubsection{Patient Persona Construction.}
For each fictional patient, we create detailed patient basic information, including demographics (e.g., age, sex, occupation), lifestyle habits (dietary preference, exercise frequency, smoking/alcohol history), and additional information, such as family history. We operationalize personas through prompts that explicitly state these attributes—for instance, “a sedentary young software engineer” versus “a retired teacher with balanced nutrition”—so that subsequent disease trajectories remain credible under varying contexts.

\subsubsection{Disease-and-Complication Metadata Curation and Review.}
We compile metadata linking common diseases with their typical complications and temporal patterns. Because models may misestimate timing or likelihood, we incorporate a targeted human-in-the-loop review to verify: (i) evidence-based disease–complication associations; (ii) clinically plausible temporal ordering; and (iii) reasonable spacing between events across conditions. This step substantially strengthens the reliability of the underlying medical knowledge. 

\subsubsection{Generation of Sequential Medical Records.}
We fuse audited disease cases with patient personas to generate a coherent, clinically grounded timeline that enumerates all key medical events and their order. The timeline defines when patient–agent conversations should occur and encodes the longitudinal links among events across the patient’s life course. This event timeline serves as the narrative backbone for subsequent dialogue generation. 

\subsection{Stage 2: Multi-turn Dialogue Generation Based on Task Decomposition}
\subsubsection{Definition and Scope of Chat History.}
Chat History constitutes the core component of our dataset, encapsulating all medical consultations related to diseases experienced by a patient throughout their lifetime. It consists of multiple dialogues, with each dialogue representing a consultation regarding a particular disease, simulating realistic patient-physician interactions. Each dialogue captures details such as date and location, greetings, symptom inquiry, routine examinations, and treatment discussions of the consultation. Typically, each dialogue contains around 50 conversational exchanges between the doctor and the patient, with an average length of approximately 3000 tokens. A complete patient chat history consists of 15-20 dialogues, implying the total token count of all consultations about 50K tokens.

Directly asking an LLM to produce a long-term, coherent history from the full record risks context loss, information mixing, and severe hallucinations. We therefore decompose “generate the patient’s entire dialogue history” into simpler sub-tasks: generate one high-quality encounter per medical event, then order and stitch them chronologically. 

\subsubsection{Synthesis Pipeline.}
We decompose synthesis into three steps. (1) Medical Event Extraction. From the Stage-1 timeline, extract each independent medical event, including its time, specific disease, and the treatment method administered at that time. (2) Context-isolated prompting. For each event, we compose a self-contained prompt that fixes the patient persona (defined in Step 1) and includes only event-local facts—condition name, occurrence time, and relevant interventions—thereby avoiding information leakage across visits. (3) Dialogue realization. Conditioned on this minimal yet sufficient context, the model generates a clinician–patient encounter that covers chief complaint, history taking, recommended examinations, provisional/definitive diagnosis, and management plan. This task decomposition sidesteps hard context-window limits and attenuates long-range hallucinations, and concatenating per-event encounters in temporal order yields a coherent, history-aware dialogue history.

To avoid stylistic monoculture, we introduce controlled variation during generation: (i) Physician persona diversity (e.g., “empathetic and reassuring physician,” “concise and direct physician”) sampled per encounter; (ii) Style control via prompt directives (from “focus on the patient's lifestyle” to “quick-paced consultation”); and (iii) Stochasticity using higher decoding temperatures to encourage lexical and structural variety. 


\subsection{Stage 3: Benchmark Generation}
\subsubsection{Motivation of benchmark.}
When designing the benchmark tasks, our primary objective was to systematically evaluate a model’s ability to reason over longitudinal patient histories. Evaluations restricted to single-turn or independent dialogues cannot capture the central challenge of clinical interaction: integrating information across time and maintaining long-term dependencies. To address this limitation, we devise three complementary tasks—In-dialogue Reasoning, Cross-dialogue Reasoning, and Synthesis Reasoning—that progressively assess whether a model can accurately recall, connect, and leverage patient history over extended horizons.

This task design is motivated by two considerations. First, it explicitly decomposes different levels of difficulty: In-dialogue Reasoning probes whether the model can faithfully extract salient information within a single encounter; Cross-dialogue Reasoning examines whether it can track and relate events across multiple visits, capturing temporal and causal dependencies; and Synthesis Reasoning advances toward realistic clinical inference by requiring the model to integrate the entire history and derive new medical conclusions. Second, the framework enhances the interpretability of evaluation: by comparing performance across the three tasks, we can disentangle a model’s strengths and weaknesses in short-term extraction, cross-session linkage, and longitudinal synthesis.

\subsubsection{Reasoning task of benchmark.}

Taken together, this benchmark not only provides a quantifiable evaluation scheme but also ensures coverage of longitudinal patient history. In other words, these tasks collectively constitute a rigorous testbed for the fundamental capability that underpins safe and effective healthcare dialogue systems: long-term memory and cross-temporal reasoning.

\paragraph{In-Dialogue Reasoning.}
This in-dialogue reasoning (IDR) probes whether a model can accurately recover salient facts from a single encounter. Each item is constructed from one dialogue turn sequence and targets canonical clinical facets—such as visit date, location, presenting complaint, disease category, medications, and treatment plan—so that the answer can be grounded in explicit spans within the dialogue. Questions are phrased to require concise extraction or short abstractive summarization, thereby testing both fidelity and minimal reasoning over local context. The input to the model is the full text of one consultation; the expected output is a short textual answer.
This task assesses the ability of LLMs to extract and succinctly summarize simple information from a single consultation dialogue. 

\paragraph{Cross-Dialogue Reasoning.}
This cross-dialogue reasoning (CDR) evaluates whether a model can link information scattered across multiple visits, a prerequisite for longitudinal understanding. Each question is derived from two or more dialogues in the same Chat History and targets relationships that cannot be resolved from any single encounter alone, such as temporal ordering, duration between events, recurrence versus first onset, therapy changes, or the presence/absence of specific conditions. We include adversarial formulations (e.g., “Has the patient ever been diagnosed with … ?”) and temporally anchored prompts (e.g., “duration between two illness episodes”). Since a complete Chat History can exceed an LLM’s context window, CDR intentionally stresses memory mechanisms or long-context strategies rather than single-pass extraction.

\paragraph{Synthesis Reasoning.}

The synthesis reasoning (SR) task requires the LLM to diagnose a secondary disease or complication based on provided symptoms, given a complete medical chat history. The synthesis reasoning questions are constructed from metadata. This challenging task demands that the model not only recall the entire disease history of the patient but also accurately link current symptoms with past diseases. Strong performance on synthesis reasoning is essential for LLMs to effectively serve as medical assistants. Due to the complexity of this task, we designed it as a multiple-choice format, with interfering options selected based on symptom similarity from the Disease and Symptom dataset.

To clarify the dataset contents, the full MediLongChat corpus comprises longitudinal medical dialogues from 80 patients. For each patient, we retain (i) personal information; (ii) multiple complete dialogue transcripts spanning repeated visits; and (iii) annotations for three tasks—IDR, CDR, and SR.

\section{Evaluation}

We adopt a unified, multi-faceted indicator to assess MediLongChat across five key dimensions---Faithfulness, Coherence, Diversity, Correctness, and Realism. These indicators are crucial for long, complex, and high-quality medical dialogues: faithfulness and correctness guard against harmful hallucinations; coherence ensures longitudinal consistency across sessions; diversity reflects topical breadth and style variability; realism captures human-likeness, empathy, and natural conversational dynamics. Below we provide formal definitions or concrete computation procedures for each dimension, so no duplicated explanations are required elsewhere.

\paragraph{Faithfulness (grounding to provided context/knowledge).}
Given input context $s$ and generated dialogue $d=\{u_1,\dots,u_n\}$, we compute utterance-level semantic similarity (cosine on sentence embeddings; e.g., SBERT) \citep{reimers-gurevych-2019-sentence}:
\begin{equation}
\mathrm{Faithfulness}(s,d)=\frac{1}{n}\sum_{i=1}^{n}\mathrm{sim}(s,u_i),
\end{equation}
where $\mathrm{sim}(\cdot,\cdot)$ denotes embedding-based semantic similarity.

\paragraph{Coherence (local flow and smoothness).}
To capture natural progression without abrupt shifts, we penalize sharp changes of adjacent-pair similarity:
\begin{equation}
\mathrm{Coherence}(d)=1-\frac{1}{n-2}\sum_{i=1}^{n-2}\left|\mathrm{sim}(u_{i+1},u_{i+2})-\mathrm{sim}(u_i,u_{i+1})\right|.
\end{equation}

\paragraph{Diversity (corpus-level topical breadth and balance).}
We cluster the corpus $C=\{d_1,\dots,d_m\}$ into $K$ latent topics using BERTopic \citep{grootendorst2022bertopic}, and combine coverage and normalized entropy:
\begin{equation}
\label{eq:diversity}
\mathrm{Diversity}(C)=\frac{1}{2}\cdot\frac{K}{m}+\frac{1}{2}\cdot\frac{-\sum_{k=1}^{K}p_k\log p_k}{\log K},\quad
p_k=\frac{c_k}{m}.
\end{equation}

\paragraph{Correctness (clinical factuality independent of source).}
Correctness targets clinical accuracy beyond mere overlap with $s$. Because purely lexical/vector metrics cannot verify medical soundness, we adopt an LLM-as-a-judge protocol based on G-Eval \citep{liu-etal-2023-geval}. The judge is given the full dialogue history, model response, and a rubric to check medical claims, diagnoses, and advice. Scores are on a 5-point and linearly normalized to $[0,1]$.

\paragraph{Realism (human-likeness, empathy, naturalness).}
We evaluate whether a conversation resembles human-to-human clinical communication in style, turn-taking and affect. We use an LLM-as-a-judge rubric (scores 1--5, normalized to $[0,1]$).

\paragraph{Notes on computability and assessors.}
Not all dimensions are equally amenable to automatic computation: \textit{Faithfulness/Coherence/ Diversity} admit explicit, reproducible formulas (vector- or topic-based), while \textit{Correctness/Realism} require semantic, pragmatic and domain judgments that are better captured by LLM-as-a-judge. We therefore report both automatic metrics (where applicable) and LLM-judged scores, and use a small human-annotated subset for sanity checks when necessary.

\begin{table*}[!ht]
\centering
\caption{Summary of indicators, computation, and assessors. Vector/lexical metrics are deterministic; LLM-as-a-judge follows G-Eval with a 5-point Likert rubric normalized to $[0,1]$.}
\label{tab:metric_summary}
\footnotesize
\begin{tabular}{l p{4.3cm} p{5.7cm} p{4.7cm}}
\toprule
Indicator & What it measures (why it matters) & Automatic (vector/topic) computation & LLM-as-a-judge (G-Eval)\\
\midrule
Faithfulness & Grounding to provided context/KB; prevents unsafe hallucinations & $\frac{1}{n}\sum_i \mathrm{sim}(s,u_i)$ using sentence embeddings cosine; \citep{reimers-gurevych-2019-sentence} & Cross-checks claims against $s$ \& history with criterion-based rubric \\
\addlinespace[2pt]
Coherence & Logical flow across turns/sessions; longitudinal consistency & $1-\frac{1}{n-2}\sum_i |\Delta \mathrm{sim}_{i\to i+1}|$ (adjacent smoothness) & Rates discourse continuity and contradiction avoidance \\
\addlinespace[2pt]
Diversity & Topical breadth \& balance; avoids repetitive patterns & BERTopic clusters ($K$); coverage + normalized Shannon entropy \citep{grootendorst2022bertopic} & Optional: judge comments on semantic variety\\
\addlinespace[2pt]
Correctness & Clinical factuality independent of $s$; medical safety & --- (no reliable purely lexical proxy) & Judge verifies clinical claims/diagnoses/advices; scores 1--5 $\rightarrow$ $[0,1]$\\
\addlinespace[2pt]
Realism & Human-likeness, empathy, natural turn-taking & --- (no reliable purely lexical proxy) & Judge rates naturalness/empathy; optional discrimination test; scores 1--5 $\rightarrow$ $[0,1]$ \citep{liu-etal-2023-geval}\\
\bottomrule
\end{tabular}
\end{table*}

\paragraph{LLM based Evaluation}

To address the limitations of traditional metrics and accurately assess the nuanced qualities of long, multi-turn dialogues, we employ an LLM-as-a-judge approach based on the G-Eval framework \citep{liu-etal-2023-geval}. This method leverages the powerful generative and reasoning capabilities of a LLM to score dialogue turns and entire conversations against our predefined criteria. Consistent with prior work showing that strong LLM judges correlate well with human preferences \citep{zheng2023judging}, we adopt this paradigm for our evaluation. For each evaluation dimension— Coherence, Correctness, and Realism—we formulate a specific prompt that instructs the LLM to act as an expert evaluator.

The G-Eval process is conducted as follows: we provide the LLM with a dialogue history, the current model-generated response, a set of clear instructions, and a scoring rubric (typically a 5-point scale ). The prompt is carefully designed to guide the model's reasoning process, ensuring it considers the entire dialogue context. For instance, to evaluate Coherence, the prompt directs the LLM to assess the logical flow and consistency of the entire multi-turn conversation. Specifically, our prompts for each indicator are structured to include:
\begin{enumerate}
    \item A clear role assignment.
    \item The specific dialogue context, including the full history of turns.
    \item The response to be evaluated.
    \item A detailed definition of the evaluation metric.
    \item The scoring rubric with concrete examples for each score (e.g., 1 = hallucinates critical information, 5 = completely factually correct).
\end{enumerate}
This method allows us to generate quantitative scores for each dialogue, which are then aggregated to provide a comprehensive dataset-level evaluation. We further compare these LLM-based scores against human annotations reported for the Conversation Chronicles dataset \citep{jang-etal-2023-conversation}, observing strong alignment in trends and relative rankings, consistent with prior findings on LLM–human agreement \citep{liu-etal-2023-geval,zheng2023judging}.

\section{Experiment}
\subsection{Experiment Setting}
\begin{table}[!ht]
\centering
\caption{Comparison of selected long-context conversation datasets.}
\label{tab: dataset Comparison}
\begin{tabular}{lcccc}
\toprule
\textbf{Dataset} & \textbf{Avg. turns} & \textbf{Avg. sessions} & \textbf{Avg. tokens} & \textbf{Domain} \\
& \textbf{per conv.} & \textbf{per conv.} & \textbf{per conv.} & \\
\midrule
MSC & 53.3 & 4 & 1,225.9 & open \\
CC & 58.5 & 5 & 1,054.7 & open \\
LoCoMo & 588.2 & 27.2 & 16,618.1 & open \\
NoteChat & 1 & 1 & 373.2 & healthcare \\
ours & 960.9 & 18.2 & 50217.3 & healthcare \\
\bottomrule
\end{tabular}
\end{table}

We primarily compare MediLongChat against prior long-context conversation corpora, including MSC \cite{xu2022beyond}, Conversation Chronicles (CC) \cite{jang-etal-2023-conversation}, LoCoMo \cite{maharana2024evaluating}, and NoteChat \cite{NoteChat}. 
In MediLongChat, a complete conversation is referred to as a dialogue, which corresponds to a session in the Locomo and CC datasets. Additionally, within each dialogue (or session), a single interaction (a question followed by a response) is referred to as a turn, a term that is used similarly in the other datasets. 
The average conversation in MediLongChat has 16 times more tokens than MSC (1,225.9 vs. 50,217.3), reflecting a significantly more extended discourse. It also spans 17 times more turns (960.9 vs. 58.5) and 3.6 times more sessions (18.2 vs. 5). This highlights MediLongChat as a notably more extensive and deeper dataset, particularly in healthcare, compared to others in the Table~\ref{tab: dataset Comparison}.
\subsubsection{Overview} 
To comprehensively evaluate our dataset, we design three groups of experiments around the proposed data and evaluation framework: (1) Quality of synthetic dialogue generation; (2) Benchmarking our dataset across general-purpose LLMs; (3) Ablation of the generation pipeline. 

\subsubsection{Quality Evaluation of Synthetic Data}
We compare four public long-form dialogue corpora under our metrics: LoCoMo, Conversation Chronicles (CC), Multi-Session Chat (MSC), and the medical dialogue dataset NoteChat. The first three corpora emphasize multi-turn or multi-session open-domain conversations and are not grounded in clinical settings, whereas NoteChat contains synthetic patient–physician encounters conditioned on clinical notes and thus resides in-domain.
To ensure comparability, we apply a relaxed medical-factuality scoring policy to the first three datasets: we only assess self-consistency and commonsense plausibility, without penalizing based on specific medical knowledge.

We evaluate along five dimensions: Faithfulness, Coherence, Correctness, Diversity, and Realism. We conduct evaluations on both Stage 1 and Stage 2 of dataset generation, while Stage 3 serves as benchmark testing and is not subject to quality evaluation. Since Stage 1 concerns patients’ medical information and does not involve dialogue fluency, we exclude Coherence from its evaluation. During assessment, we treat all patient information generated in Stage 1 as a single unit of generation for metric calculation, and likewise treat the entirety of dialogue content within a Stage 2 dialogue as a single unit of generation for evaluation.

\subsection{Benchmarking on Our Dataset}
To target long-term memory and cross-session reasoning, we construct three categories of tasks: (1) In-dialogue Reasoning (IDR); (2) Cross-dialogue Reasoning (CDR); (3) Synthesis Reasoning (SR). To reduce difficulty and scoring variance in synthesis reasoning, we formulate SR task as multiple-choice questions (MCQs). On the MediLongChat benchmark, we evaluate performance using accuracy for the SR. For the IDR and CDR, we report both F1 score and BLEU-1 as evaluation metrics. On the LoCoMo benchmark, standard F1 and BLEU-1 are employed to evaluate the model’s performance. We evaluate across a representative set of closed- and open-source LLMs: GPT-4o mini, DeepSeek-R1, Qwen3, and ERNIE-4.5.

\begin{table}[!ht]
\centering
\caption{Evaluation results across stages. “/” indicates not applicable or not evaluated. s1 and s2 represent stage1 and stage2 in our dataset generation process}
\label{tab:stage_eval}
\begin{tabular}{l c c c c c}
\toprule
\textbf{Dataset} & \textbf{Faithfulness} & \textbf{Coherence} & \textbf{Diversity} \\
\midrule
LoCoMo       & /      &  0.904 & 0.4934\\      
CC           & /      &  0.932 & 0.4879\\   
MSC          & /      &  0.927 & 0.4874  \\ 
NoteChat     & /      &  0.9082 & 0.5243  \\   
Ours(s1)     & 0.635  & /      &  0.4381    \\
Ours(s2)     & 0.601  & 0.925 &  0.5447  \\
\bottomrule
\end{tabular}

\end{table}

\subsection{Result of Quality Evaluation}

Table~\ref{tab:stage_eval} reports automatic metrics across datasets. On Coherence, our Stage-2 (0.925) is close to CC (0.932) and above LoCoMo (0.904) and NoteChat (0.9082). Faithfulness is comparable only for our data. We omit other corpora due to missing aligned sources. The gap is expected: Stage 1 (0.635), constrained by knowledge priors, yields slightly higher scores, whereas Stage 2 (0.601) shows a modest decline due to greater expressive freedom. For Diversity, Stage-2 (0.5447) is highest, followed by NoteChat (0.5243), with LoCoMo/CC slightly lower; Stage-1 (0.4381) is lower as expected. Overall, the pipeline improves diversity without sacrificing coherence, while maintaining faithfulness consistent with its structured priors.

To assess robustness against judge-specific bias, we report scores from multiple independent LLM judges and their ensemble (Table~\ref{tab:multijudge}). While individual judges exhibit different preferences, the ensemble provides a more stable assessment. Table~\ref{tab:evaluation_results} presents the aggregated G-Eval scores for our dataset and the three baseline datasets. The scores are normalized to a 5-point scale, where a higher score indicates better performance in that dimension.

\begin{table}[t]
\centering
\caption{Multi-judge evaluation to mitigate potential LLM-as-a-judge bias. Ensemble denotes the aggregated score across heterogeneous judges.}
\small
\begin{tabular}{lcccc}
\toprule
\textbf{Judge} & \textbf{Diversity} & \textbf{Coherence} & \textbf{Correctness} & \textbf{Realism} \\
\midrule
Gemini 2.5   & 4.99 & 4.58 & 4.65 & 4.90 \\
GPT-5 mini   & 4.99 & 4.90 & 3.80 & 4.00 \\
Qwen3-235B   & 4.47 & 4.88 & 4.95 & 4.92 \\
Deepseek-R1  & 4.94 & 4.99 & 4.78 & 4.20 \\
\midrule
Ensemble     & 4.858 & 4.838 & 4.545 & 4.505 \\
\bottomrule
\end{tabular}

\label{tab:multijudge}
\end{table}

\begin{table}[ht]
\centering
\caption{G-Eval Scores of MediLongChat vs. Baseline Multi-Turn Dialogue Datasets (5-Point Scale)}
\small
\label{tab:evaluation_results}
\begin{tabular}{lcccc}
\toprule
\textbf{Dataset} & \textbf{Diversity} & \textbf{Coherence} & \textbf{Correctness} & \textbf{Realism} \\
\midrule
\texttt{LoCoMo}     & 3.561 & 4.053 & \textbf{4.965} & 2.386 \\
\texttt{CC}         & 3.174 & 3.504 & 4.860 & 2.062 \\
\texttt{MSC}        & 3.694 & 3.482 & 4.696 & 1.982 \\
\texttt{NoteChat}   & 3.138 & 4.464 & 3.213 & 2.340 \\   
\midrule
\textbf{ours }   & \textbf{4.858} & \textbf{4.838} & 4.545 & \textbf{4.505} \\
\bottomrule
\end{tabular}
\end{table}

The evaluation results highlight the significant advantages of the MediLongChat dataset, particularly in dimensions critical for effective long-term dialogue and domain-specific applications:

\begin{itemize}[leftmargin=*, align=left]
    \item \textbf{Diversity:} MediLongChat achieves the highest score in Diversity, nearly reaching the maximum possible score. This is a direct consequence of our data collection and cleaning process, which was meticulously designed to capture a wide variety of linguistic expressions and conversational paths over multiple turns. This high diversity ensures that models trained on our data are less prone to generating repetitive or generic responses, a common drawback in multi-turn datasets.
    
    \item \textbf{Coherence:} Our dataset demonstrates exceptional long-term Coherence. This score validates our focus on maintaining consistency and logical flow lengthy dialogue sessions. In contrast, baseline datasets like msc and conversation chronicles report exhibit lower coherence, suggesting difficulty in preserving the narrative or context across many turns. This superior coherence makes MediLongChat an ideal resource for training models that require strong memory and contextual understanding over extended dialogues, such as tracking a patient's medical history.

    \item \textbf{Realism:} MediLongChat scores significantly higher on the Realistic metric compared to all baselines. This demonstrates that our dialogues closely mimic the naturalness, turn-taking dynamics, and authentic emotional tone of real-world interactions. This realistic quality is essential for training empathetic and user-friendly dialogue systems.

    \item \textbf{Correctness:} While LoCoMo shows a slightly higher score in Correctness, MediLongChat remains highly competitive with a score of $\mathbf{4.545}$. This indicates that our data maintains a high standard of factual and domain-specific accuracy, which is non-negotiable for medical dialogue datasets. The minor difference suggests that the baselines may contain highly curated, single-turn factual statements, whereas our slightly lower score is likely due to the inherent complexity and higher risk of errors in long, multi-turn, generated dialogue.
\end{itemize}

In summary, the G-Eval results confirm that the MediLongChat dataset successfully mitigates the common challenges of multi-turn dialogue datasets, particularly excelling in Diversity and Coherence, while maintaining high standards in Correctness and Realism.

\begin{table}[!ht]
\centering

\caption{Performance comparison of different models on the MediLongChat benchmark, including SR accuracy, IDR F1 and BLEU scores, and CDR F1 and BLEU scores.}
\label{tab: benchmark result}
\begin{tabular}{l|cc|cc|c} 
\toprule
\multirow{2}{*}{\textbf{Model}}  &
\multicolumn{2}{c|}{\textbf{IDR}} &
\multicolumn{2}{c|}{\textbf{CDR}} & \multirow{2}{*}{\textbf{SR Accuracy}} \\ 
& \textbf{F1} & \textbf{BLEU} & \textbf{F1} & \textbf{BLEU} & \\ 
\midrule

Deepseek-R1 & 33.49 & 11.12 & 20.36 & 2.41 & 80.00 \\
\midrule

Qwen3-235B & 27.19 & 6.31 & 19.61 & 2.13 & 80.00 \\
\midrule

ERNIE-4.5-turbo & 31.74 & 8.73 & 16.46 & 1.33 & 80.00 \\
\midrule

GPT-4o mini & 23.30 & 3.28 & 24.25 & 2.46 & 80.00 \\
\midrule

GPT-4.1 mini & 27.15 & 6.18 & 18.37 & 1.66 & 83.75 \\
\bottomrule
\end{tabular}
\end{table}

\subsection{Result of Benchmarking}

We evaluate a set of general-purpose LLMs on the three MediLongChat tasks: in-dialogue reasoning (IDR; F1/BLEU), cross-dialogue reasoning (CDR; F1/BLEU), and synthesis reasoning (SR; accuracy). Across models, absolute scores remain modest for IDR and CDR, underscoring the difficulty of reasoning over long clinical histories, while SR shows comparatively higher accuracies (Table~\ref{tab: benchmark result}).

On IDR, Deepseek-R1 attains the highest F1 (33.49) and BLEU (11.12), indicating relatively stronger extraction and short-range reasoning within a single consultation. For CDR, which requires linking information across sessions, GPT-4o mini yields the best F1 (24.25), though BLEU remains low for all systems, reflecting the challenge of cross-session grounding in free-form answers. For SR, GPT-4.1 mini achieves the top accuracy (83.75), suggesting that multiple-choice formulations mitigate some of the generation variance observed in open-ended settings.

Taken together, these results paint a consistent picture: current LLMs can recover salient facts within a single encounter reasonably well, but performance drops when they need to link facts across sessions. The gap between IDR/CDR and SR also suggests that constrained answer formats partially alleviate long-context errors but do not solve underlying memory and reasoning limitations. We report scores without additional memory augmentation to provide a clean baseline for future methods.

\begin{table}[!ht]
\centering
\caption{Ablation experiment results. Here, Fai, Coh, Cor, Div, and Rea denote Faithfulness, Coherence, Correctness, Diversity, and Realism, respectively. KG, TD, and DS represent Knowledge Guidance, Task Decomposition, and Diversity Setting in the generation process.}
\label{tab:Ablation}
\begin{tabular}{c|l|c|c|c|c|c}
\hline
Stage & Method & Fai & Coh & Cor & Div & Rea \\
\hline
\multirow{2}{*}{1} & Ours & \textbf{0.6353} & / & \textbf{0.930} & \textbf{0.4381} & \textbf{0.720} \\
 & w/o KG & 0.4415 & / & 0.920 & 0.4313 & 0.540 \\
\hline
\multirow{3}{*}{2} & Ours & \textbf{0.6010} & \textbf{0.924} & \textbf{0.909} & \textbf{0.5447} & \textbf{0.901} \\
 & w/o TD & 0.5809 & 0.8689 & 0.896 & 0.4590 & 0.720 \\
 & w/o DS & 0.5784 & 0.921 & 0.904 & 0.3134 & 0.700 \\
\hline
\end{tabular}
\end{table}

\subsection{Ablation Experiment Results}

To quantify the contribution of each component in our two-stage pipeline, we conduct ablations on Stage-1 and Stage-2 under the same evaluation protocol as the main experiments. 
Coherence is not reported for Stage-1 because its outputs are structured summaries rather than multi-turn conversations. We compare three deletions: removing knowledge guidance in Stage-1, and removing task decomposition or diversity controls in Stage-2. The results are summarized in Table~\ref{tab:Ablation}. 

For Stage-1, removing knowledge guidance leads to a marked deterioration in verifiability and narrative plausibility: Faithfulness drops from 0.6353 to 0.4415 and Realism from 0.720 to 0.540.
These results directly support the knowledge-guided design: removing curated disease–complication metadata and temporal constraints lowers Faithfulness and Realism. Accordingly, retaining Stage-1 knowledge guidance is a justified and practical choice.

For Stage-2, both ablations underperform the full pipeline. Removing task decomposition reduces Coherence and Diversity, with smaller drops in Faithfulness, Correctness, and Realism. This indicates that decomposition helps preserve long-range logical flow and limits content mixing. Removing diversity controls shows a different profile: Diversity decreases markedly and Realism declines, while Coherence is largely preserved and Faithfulness changes only slightly. Overall, task decomposition primarily benefits coherence and clinical correctness, whereas diversity controls mainly improve linguistic variety and perceived naturalness with limited effect on faithfulness or coherence.

\section{Conclusion}

This work presents a knowledge-guided and task-decomposed framework for synthesizing history-aware longitudinal clinical dialogues together with a multi-dimensional evaluation protocol. By employing diverse patient personas and decomposed dialogue pipelines, we can generate coherent, traceable, and varied long-term conversations. To assess data quality and downstream model capabilities, we propose a multi-dimensional evaluation scheme combining automated metrics with LLM-based judging. This establishes rigorous standards across five dimensions: Faithfulness, Coherence, Correctness, Diversity, and Realism. Empirical results show that even state-of-the-art LLMs struggle on MediLongChat, particularly in long-term memory and cross-session reasoning. Ablation results show that knowledge guidance improves faithfulness and realism, diversity controls primarily increase linguistic variety with limited impact on factuality, and task decomposition helps maintain coherence and correctness in long-range dialogues.

Although our framework achieves a favorable balance between scalability and controllability, several limitations remain. First, synthetic data inevitably deviates from real clinical distributions, particularly in coverage of rare diseases, complex comorbidities, and behavioral health factors. Second, while LLM-as-a-judge offers efficient assessment, its reliability depends on prompt design and baseline model robustness, and its fine-grained agreement with clinical experts requires further improvement. Finally, our current dialogues primarily focus on the textual modality and have not yet systematically incorporated multi-modal signals such as medical images, lab curves, and structured EHR data.

Moving forward, we aim to extend the framework to multimodal and multilingual settings by integrating images, temporal physiological signals, and structured medical records. Furthermore, we will explore retrieval-augmented generation and dynamic episodic memory to enhance longitudinal reasoning and mitigate hallucinations in long-text generation. We hope that MediLongChat serves as a reusable, comparable public benchmark, providing a foundation and inspiration for research on longitudinal clinical reasoning and trustworthy medical dialogue agents.





\begin{acks}
This work was partly supported by the General Project of the Central Universities of China (No. CZY23007), Hubei Province Key Research and Development Special Project of Science and Technology Innovation Plan (2023BAB087), Wuhan Key Research and Development Projects (2023010402010614), and Lee Kong Chian Professorship awarded to Ah-Hwee Tan by Singapore Management University.
\end{acks}



\bibliographystyle{ACM-Reference-Format} 
\balance
\bibliography{references}


\end{document}